\documentclass{article}

\usepackage[english]{babel}

\usepackage[letterpaper,top=2cm,bottom=2cm,left=3cm,right=3cm,marginparwidth=1.75cm]{geometry}

\usepackage{amsmath}
\usepackage{amsfonts}
\usepackage{bm}
\usepackage{graphicx}
\usepackage{amsthm}
\usepackage{microtype}
\usepackage{subfigure}
\usepackage{caption}
\usepackage{booktabs}
\usepackage[normalem]{ulem}

\usepackage[colorlinks=true, allcolors=blue]{hyperref}

\newtheorem{definition}{Definition}

\title{Transfer Operator Learning with Fusion Frame}

\author{Haoyang Jiang, Yongzhi Qu\\
Mechanical Engineering, The University of Utah, 1495 E 100 S, Salt Lake City, 84112}

\begin{document}
\maketitle
\begin{abstract}
The challenge of applying learned knowledge from one domain to solve problems in another related but distinct domain, known as transfer learning, is fundamental in operator learning models that solve Partial Differential Equations (PDEs). These current models often struggle with generalization across different tasks and datasets, limiting their applicability in diverse scientific and engineering disciplines. This work presents a novel framework that enhances the transfer learning capabilities of operator learning models for solving Partial Differential Equations (PDEs) through the integration of fusion frame theory with the Proper Orthogonal Decomposition (POD)-enhanced Deep Operator Network (DeepONet). We introduce an innovative architecture that combines fusion frames with POD-DeepONet, demonstrating superior performance across various PDEs in our experimental analysis. Our framework addresses the critical challenge of transfer learning in operator learning models, paving the way for adaptable and efficient solutions across a wide range of scientific and engineering applications.
\end{abstract}

\section{Introduction}
\subsection{Motivation}

The rapid advancement in scientific machine learning methods has shown a new era in the field of mathematical modeling and simulation, particularly in solving Partial Differential Equations (PDEs) that are ubiquitous across various scientific and engineering disciplines. Operator learning emerges as a potent framework for understanding mappings between infinite-dimensional function spaces, offering promising avenues for tackling PDEs in a data-driven manner\cite{kovachki2023neural}. Despite its potential, the application of operator learning models is often hampered by a critical challenge: the limited ability to transfer knowledge gained from one task (source domain) to a related but distinct task (target domain). This limitation not only restricts the adaptability of models to new, unseen datasets but also reduces their efficiency in solving novel equations not encountered during the initial training phase.


The importance of overcoming these challenges is crucial, as it greatly affects the usefulness of models in real-world settings where conditions and needs can differ greatly. In fields such as fluid dynamics, material science, climate modeling, and biomedical engineering, the ability to quickly adapt and use knowledge in different situations is extremely valuable. Although current neural operator methods are good at handling specific tasks, they often struggle to adapt quickly without a lot of retraining. This is a major problem, especially in situations where there are limited computational resources or it is hard to gather data.


Recognizing these gaps, our research is propelled by a motivation to bridge the divide between the potential of operator learning models and their current limitations in transferability. By integrating advanced mathematical frameworks such as fusion frame theory with neural network architectures, we aim to unlock new possibilities in operator learning, enabling models that are not only robust and accurate but also highly adaptable across diverse problem settings. The endeavor to enhance transfer learning capabilities in operator learning models is more than an academic pursuit, it is a crucial step toward realizing real-world applications.

\subsection{Contributions}

This work introduces a novel fusion frame-based framework for enhancing the transfer learning capabilities of operator learning models, specifically for solving partial differential equations (PDEs) across various domains. Our contributions address several key challenges in the field, offering both theoretical advancements and practical implications:

\begin{itemize}
    \item \textbf{Fusion Frame-enhanced POD-DeepONet Architecture:} We propose an architecture that incorporates fusion frame theory into the Proper Orthogonal Decomposition (POD) enhanced Deep Operator Network (DeepONet). This architecture not only broadens the representational capacity of neural networks in capturing the mappings between infinite-dimensional spaces but also significantly improves their ability to generalize across different tasks and datasets.

    \item \textbf{Robust Transfer Learning Methodology:} Our framework establishes a robust methodology for transfer learning in operator learning models. By leveraging the principles of fusion frames, we facilitate a more effective and efficient knowledge transfer process, enabling the models to perform adeptly on new equations and unseen datasets with minimal retraining.
\end{itemize}

These contributions not only address the immediate challenge of enhancing transfer learning in operator learning models but also provide a solid foundation for future studies, our work paves the way for the development of more sophisticated and adaptable models, capable of solving complex PDEs across a broad spectrum of scientific and engineering disciplines.

\section{Problem Setup}

In operator learning, the goal is to approximate an operator $G: X \rightarrow Y$ that maps a set of input functions $x \in X$ to a set of solution functions $y \in Y$. This mapping is particularly useful in solving Partial Differential Equations (PDEs), where $x$ could represent coefficients, forcing terms, or boundary conditions parameterizing the PDE, and $y$ represents the PDE solution. In our work, let $DS = \{(x^S_i, y^S_i)\}_{i=1}^{N_S}$ be the source dataset and $DT = \{(x^T_i, y^T_i)\}_{i=1}^{N_T}$ be the target dataset, where $x^S_i, x^T_i \in X$ and $y^S_i, y^T_i \in Y$. The objective is to learn the operator $G$ from $DS$ that can generalize well to $DT$.

To address the challenge of transfer learning in operator learning, we consider two scenarios:

\begin{enumerate}
    \item \textbf{Distribution Shift in Input Data}: The distribution of input functions shifts from the source domain $X^S$ to the target domain $X^T$, while the operator $G$ that maps inputs to outputs remains consistent. The objective is to adapt the operator learned from $DS$ to perform accurately on $DT$.
    
    \item \textbf{Variation in Specific PDE Terms}: The PDE in the source domain is defined as $L[u(x); x^S] = 0$, while in the target domain, it changes to $L[u(x); x^T] = f(x)$, where $f(x)$ represents the modified term, such as a new forcing function. The challenge is to adapt the learned mapping $G$ to accurately predict the solutions $y^T$ in the target domain under these modified conditions.
\end{enumerate}

For both scenarios, the source and target datasets consist of pairs of input functions and corresponding solution functions. The goal is to leverage the operator learned from $DS$ to predict $y^T$ from $x^T$ with high accuracy. This involves effectively addressing both the shift in the distribution of the input functions and changes in specific terms within the PDEs.

To achieve this, we define the following optimization problem for transfer operator learning:


\[
\min_{\theta} \frac{1}{N_T} \sum_{i=1}^{N_T} \| G_{\theta}(x^T_i) - y^T_i \|^2 + \lambda \mathcal{R}(\theta)
\]

where:
\begin{itemize}
    \item $G_{\theta}$ is the parameterized operator with parameters $\theta$.
    \item $x^T_i$ and $y^T_i$ are the input and solution functions from the target dataset.
    \item $\mathcal{R}(\theta)$ is a regularization term to prevent overfitting.
    \item $\lambda$ is regularization coefficient.
        
\end{itemize}

The aim is to find the optimal parameters $\theta$ that minimize the discrepancy between the predicted and actual solution functions in the target dataset while ensuring the model generalizes well to new, unseen data and effectively transfers knowledge from the source domain to the target domain. By solving this optimization problem, we enhance the transferability and robustness of the operator learning model, making it adaptable to various PDE scenarios across different scientific and engineering domains.

\section{Preliminaries}

Operator learning, a key concept in understanding mappings between infinite-dimensional function spaces, has shown promising results, especially in solving partial differential equations. We discuss how the application of frames in operator learning can lead to improved transferability and robustness.

\subsection{Preliminaries on Frame Theory}
Frames extend the concept of bases, providing a more flexible representation system in Hilbert spaces, allowing for redundancy and more flexible representations\cite{2024frame,2009fusion}.

\begin{definition}[Frame]
A family of elements $\{f_i\}_{i\in I}$ in a Hilbert space $H$ is called a frame if there exist constants $0 < A \leq B < \infty$ such that
\[
A\|f\|^2 \leq \sum_{i\in I} |\langle f, f_i \rangle|^2 \leq B\|f\|^2, \forall f \in H.
\]
\end{definition}

Frames in Hilbert spaces provide flexible representations of signals or functions. For a frame \(\{f_i\}_{i \in I}\) in a Hilbert space \(H\) with frame bounds \(A, B > 0\), any element \(f \in H\) can be reconstructed using the frame operator \(S\), defined by:
\[Sf = \sum_{i \in I} \langle f, f_i \rangle f_i.\]
The reconstruction formula is:
\[f = S^{-1}Sf = \sum_{i \in I} \langle f, f_i \rangle S^{-1}f_i,\]
where \(\{S^{-1}f_i\}_{i \in I}\) forms the dual frame.

This definition ensures that frames can provide stable, yet overcomplete, representations of signals or functions, offering a robust platform for various mathematical operations, including operator learning.

Fusion frames widen the scope of frame theory by incorporating collections of subspaces, thus offering a versatile structure for representing and processing complex data.

\begin{definition}[Fusion Frame]
A collection of weighted subspaces $\{(W_i, w_i)\}_{i\in I}$ of $\mathcal{H}$ forms a \textit{fusion frame} if there exist constants $0 < A \leq B < \infty$ such that for any $f \in \mathcal{H}$,

\[
A\|f\|^2 \leq \sum_{i\in I} w_i^2 \|P_{W_i} f\|^2 \leq B\|f\|^2,
\]
where $P_{W_i}$ denotes the orthogonal projection onto the subspace $W_i$.
\end{definition}

Fusion frames extend frames to collections of subspaces, allowing for data representation across different dimensions. A fusion frame \(\{(W_i, w_i)\}_{i \in I}\) in \(H\) with fusion frame bounds \(A, B > 0\) enables approximate reconstruction of any \(f \in H\) from its projections. The fusion frame operator \(S\) is:
\[Sf = \sum_{i \in I} w_i^2 P_{W_i}f,\]
with reconstruction given by:
\[f = S^{-1}Sf = \sum_{i \in I} w_i^2 S^{-1}P_{W_i}f.\]
This illustrates how elements can be reconstructed without requiring the subspaces \(W_i\) to be orthogonal.

Fusion frames, therefore, offer a hierarchical and fragmented representation of data, where each subspace can capture different components or features of the data, and the collective processing across these subspaces can provide a comprehensive understanding of the underlying phenomena\cite{2009fusion,2011constructingframe}.

\subsection{Frame Reconstruction in Operator Learning}
The use of frame theory in operator learning enhances the model's ability to efficiently map functions between infinite-dimensional spaces \cite{2024frame}. The reconstruction of functions from frame coefficients plays a pivotal role in this context.

Given a frame $\{f_i\}_{i \in I}$ for the input space $\mathcal{H}$, and a corresponding frame $\{g_j\}_{j \in J}$ for the output space $\mathcal{K}$, the operator learning task can be conceptualized as learning the mapping between their frame coefficients. The input function $f$ is first decomposed into its coefficients $\langle f, f_i \rangle$ with respect to the input frame. The neural operator $\hat{G}$, trained on these coefficients, predicts the coefficients of the output function in the output frame, which are then used to reconstruct the output function $g$:

\begin{equation}
    g = \sum_{j \in J} \hat{G}(\langle f, f_i \rangle) g_j,
\end{equation}

where $\hat{G}(\langle f, f_i \rangle)$ denotes the predicted coefficients for the output frame $\{g_j\}_{j \in J}$. This process underlines the importance of frames in representing and processing functions within operator learning frameworks, enabling accurate and efficient function mapping across complex domains.


\section{Method: Fusion Frame-POD-DeepONet (FF-POD-DeepONet)}

In this section, we propose a method that combines Fourier Feature Networks (FFNs) and Fusion Frames with POD-enhanced Deep Operator Network (POD-DeepONet) to improve the generalization and adaptability of operator learning models across different tasks and datasets, which better captures the features of input functions and effectively transfers knowledge.

\subsection{POD-DeepONet}

POD-DeepONet is an enhanced version of the Deep Operator Network (DeepONet), which combines Proper Orthogonal Decomposition (POD) with the standard DeepONet architecture. DeepONet consists of two networks: a branch network that processes the input functions and a trunk network that learns representations of the output space. POD is integrated into this framework to decompose functions into orthogonal modes, capturing the most energetic features of the data. By incorporating POD, the model efficiently represents the function space, leading to improved performance and generalization in operator learning tasks.

The POD modes in POD-DeepONet act as a frame for the function space, facilitating the learning process. The outputs of the POD-DeepONet can be seen as coefficients in a frame expansion, highlighting the synergy with frame theory. The subspaces spanned by POD modes can be treated as elements of a fusion frame. This allows POD-DeepONet to handle multiple scenarios or physical systems within a unified framework, demonstrating the adaptability of fusion frames in operator learning.

\subsection{Fusion Frames' Representations with POD-DeepONet}
Fusion frames consist of weighted subspaces $\{(W_i, w_i)\}$ that do not require orthogonality. The reconstruction within a fusion frame takes the form:
\begin{equation}
    f = S^{-1}\left(\sum_{i} w_i^2 P_{W_i} f\right),
\end{equation}
where $P_{W_i}$ is the orthogonal projection onto subspace $W_i$, and $S^{-1}$ is the inverse of the fusion frame operator.

Fusion Frames provides a flexible and redundant data representation. Each subspace captures different aspects of the input functions, and this hierarchical and fragmented representation enhances the model's adaptability and generalization. By allowing redundancy between subspaces, Fusion Frames improve the robustness of data processing. Integrating Fusion Frames with POD-DeepONet enhances the model's performance in complex PDE scenarios.

The integration of POD with DeepONet, combined with the principles of frame and fusion frame theory, offers a powerful approach to learning complex operators. This synergy leads to models that are both mathematically grounded and practically effective in various computational tasks.

\subsection{Subspaces based on Fourier Feature Networks}

Fourier Feature Networks (FFNs) are instrumental in constructing subspaces for fusion frames by mapping input data
to a high-dimensional spectral space. This section elaborates on the role of FFNs in defining such subspaces and their significance in the context of function space representations

\subsubsection{Fourier Feature Networks (FFNs)}

Fourier Feature Networks (FFNs) provide a transformative approach to representing and analyzing data across multiple dimensions. By leveraging the Fourier transform, FFNs project input data onto a basis that reveals its intrinsic frequency components, enabling the network to capture and process complex patterns that exist in multidimensional spaces\cite{FFN}. 

FFNs apply a series of sinusoidal transformations to input vectors, effectively enriching the feature space with high-dimensional spectral information. The core principle behind this transformation is to encode inputs using a series of sine and cosine functions, each corresponding to different frequencies, allowing the network to capture a wide range of patterns and dependencies. The transformation applied to an input vector $\bm{x} \in \mathbb{R}^d$ is defined as:

\begin{equation}
    \bm{\Phi}(\bm{x}) = \left[\cos(2\pi \bm{B}^T \bm{x}), \sin(2\pi \bm{B}^T \bm{x})\right]^T,
\end{equation}
where the matrix $\bm{B} = [\bm{b}_1, \bm{b}_2, \ldots, \bm{b}_m]$ contains the frequencies that are randomly sampled. Each vector $\bm{b}_i$ represents a specific frequency in the Fourier basis. FFNs are not limited by the dimensionality of the input space. Whether the data comes from a two-dimensional image or a higher-dimensional signal, FFNs can adaptively process and understand its fundamental structures.

FFNs have been introduced in scientific machine learning\cite{FFNPINN} for learning high-frequency information with better performance.


\subsubsection{Fusion Frame Subspace Construction}
\begin{figure}[htbp]
    \centering
    \includegraphics[trim=0cm 5cm 0cm 2cm, clip, width=\textwidth]{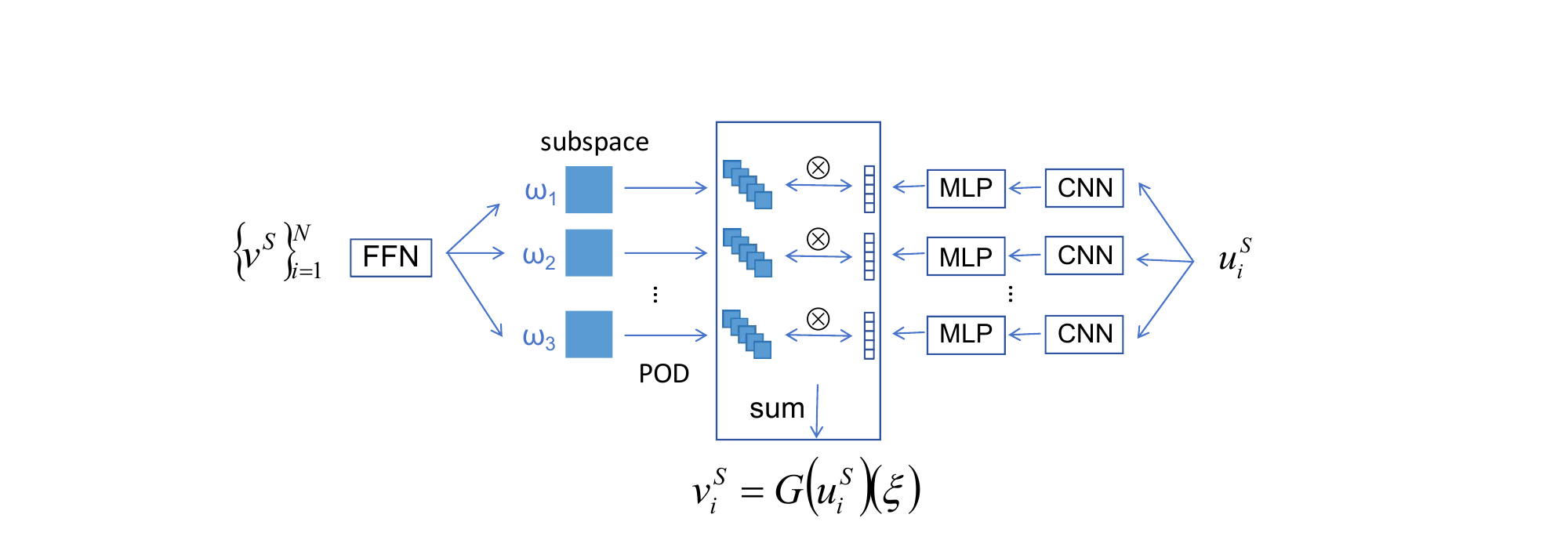}
    \caption{\label{fig:pod-ff}FF-POD-DeepONet: This figure illustrates the overall architecture of the Fusion Frame-enhanced POD-DeepONet for the source dataset. In the two-dimensional case, the input coefficients or boundary conditions \( u \) are processed by multiple neural networks, with each network generating a set of outputs. Here, \( u \) may represent certain physical parameters, such as material properties or boundary conditions. Meanwhile, the function space \( v \) is processed through Fourier Feature Networks (FFNs) to produce multiple subspaces, each associated with a learnable weight parameter \( w \). These subspaces can be considered as subspaces of the Fusion Frame, and they are processed accordingly. Each subspace \( W_i \) undergoes dimension reduction and feature extraction using POD. The outputs from these subspaces are then multiplied by the outputs from the neural networks processing \( u \), and the results are summed or averaged to obtain the final output. Finally, the query point \( \xi \) allows for querying the value of \( v \) at any point in the final result. This architecture, by combining multi-dimensional information and Fourier features, enhances the model's predictive capability and adaptability in complex physical scenarios.}
\end{figure}


\begin{figure}[htbp]
    \centering
    \includegraphics[trim=0cm 5cm 0cm 2cm, clip, width=\textwidth]{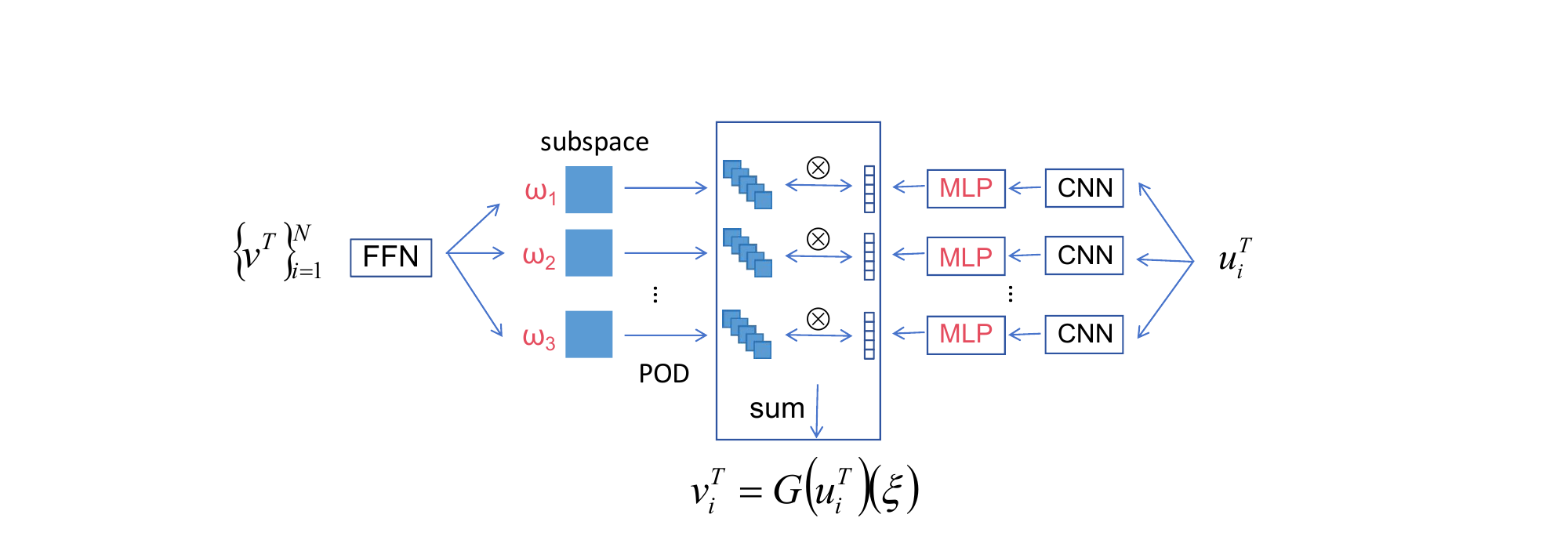}
    \caption{\label{fig:pod-ff-tl}Fusion Frame POD DeepONet in TL: In this figure for target dataset, the red parts indicate the components that need to be relearned during transfer learning. These components include the MLP processing the input function \( u \) and the learnable parameters f the function space.}
\end{figure}


Each subspace in the fusion frame is constructed using a selection of Fourier features, which capture a specific portion of the signal's frequency spectrum, we can see the construction of this whole model in Fig. \ref{fig:pod-ff}, for the transfer learning, the construction of this model is in Fig. \ref{fig:pod-ff-tl}. The collection of these subspaces encompasses the entirety of the signal's spectral content, facilitating a comprehensive representation. The subspaces, defined by their respective feature sets, can be denoted as $\{W_k\}_{k=1}^{m}$, where each $W_k$ is spanned by a subset of Fourier features:

\begin{equation}
    W_k = \text{span}\{\bm{\Phi}_k(\bm{x}) : \bm{x} \in \mathbb{R}^d\}.
\end{equation}

The integration of Fourier Feature Networks in fusion frame theory provides a powerful method for constructing subspaces that offer a rich representation of functions. This approach is particularly suited for complex signal processing tasks where capturing the full spectrum of signal behavior is crucial.






\subsubsection{Fourier Encoding}
Before input processing, data undergoes Fourier encoding to transform it into a high dimensional spectral space. This encoding enhances the model's ability to detect and utilize high-frequency data components, crucial for modeling complex physical phenomena.

Taking a single picture as an example, it is the one input of the section 5.1 example, we use FFN on this picture and get some different pictures as following Fig. \ref{fig:ffn}. 


\begin{figure*}[ht]
\centering
\includegraphics[width=1.0\textwidth]{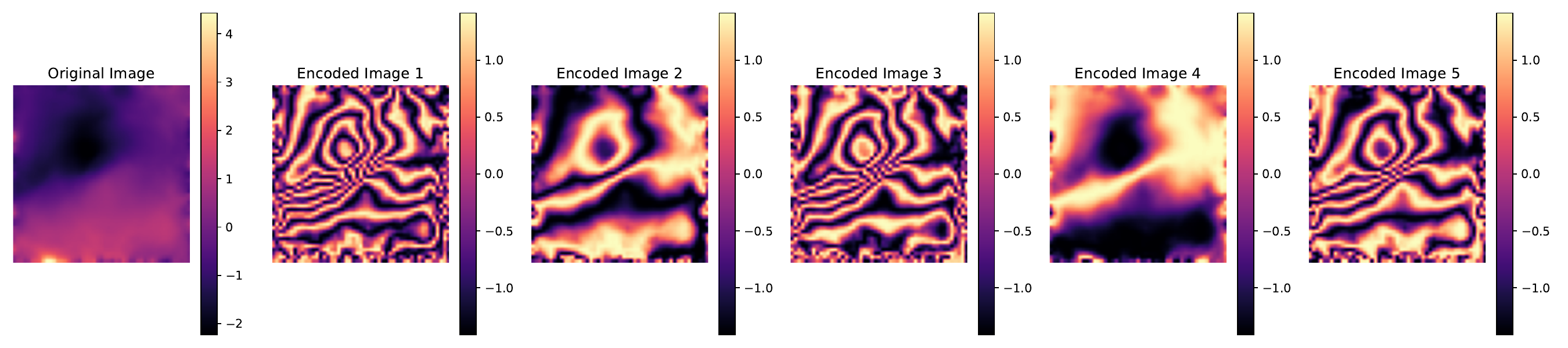} 
\caption{Fourier Feature Encoding}
\label{fig:ffn}
\end{figure*}





\section{Test Example}

Based on the above method description, to evaluate the effectiveness of our Fusion Frame POD-DeepONet framework in transfer operator learning, in two scenarios we conducted experiments comparing it with traditional POD-DeepONet, standard DeepONet, and the Fourier Neural Operator (FNO) across various PDEs. Here, we present the results of these comparisons in tabular form to highlight the performance benefits of our approach.

In the training, We use the hybrid loss function that includes a regression loss component and a conditional embedding operator discrepancy (CEOD) loss form\cite{goswami2022tldon}, and for the FNO in transfer learning, we will retrain the Projecting neural network according to the work of Zhang et al.\cite{zhang2023metano}.

\subsection{Darcy Flow}

\subsubsection{Background of Darcy flow}

Efficient simulation of Darcy flow is critical for various applications. The Darcy flow equation is formulated as:

\begin{equation}
-\nabla \cdot (a(x) \nabla u(x)) = f(x),
\end{equation}
where, \( u(x) \) represents the pressure field within the medium. \( a(x) \) denotes the permeability of the porous medium, which is a function of the spatial variable \( x \).\( f(x) \) corresponds to external forces, such as source or sink terms.

\subsubsection{Operator learning}

In the realm of operator learning, particularly for solving Partial Differential Equations (PDEs), our approach introduces a significant advancement by integrating the fusion frame theory with the Proper Orthogonal Decomposition (POD)-enhanced Deep Operator Network (DeepONet). 




Initially, we employ a Fusion Frame Network (FFN) to derive subspaces from the target dataset, effectively partitioning the input space into 20 distinct subspaces. Subsequently, POD decomposition is performed within each subspace, with an optimal selection of 80 POD modes per subspace, as empirically validated through our testing. 


To validate the effectiveness of our framework, we conducted a comprehensive comparison against the traditional POD-DeepONet, the standard DeepONet, and the Fourier Neural Operator (FNO) in the source dataset. The evaluation demonstrates our framework's superiority in terms of predictive accuracy in Table \ref{tab:darcy_source_dataset}.

\begin{table*}[h!]
\centering
\begin{tabular}{lcccc}
\toprule
Method & POD-DeepONet & DeepONet & FNO & FF POD-DeepONet \\
\midrule
MSE & 0.0624 & 0.0803 & 0.0671 & \textbf{0.0589} \\

\bottomrule
\end{tabular}
\caption{Performance comparison on the source dataset $X^S$.}
\label{tab:darcy_source_dataset}
\end{table*}

\subsubsection{Transfer Operator learning}

Furthermore, in the context of transfer operator learning, we retain the aforementioned subspaces and POD configurations while testing on the target dataset $X^T$. For the transfer operator learning, we consider two distinct scenarios, each characterized by variations in the Gaussian random field parameters or the right-hand side term of the PDEs.

In this scenario 1, we explore the impact of changes in the parameters governing the covariance of the input Gaussian random field. The target dataset \( X^T \) is sampled from a field with parameters \( \alpha = 1.2 \) and \( \tau = 1.2 \), while the source dataset \( X^S \) is sampled with \( \alpha = 2.2 \) and \( \tau = 2.2 \). The random field is defined as:
\begin{equation*}
    a(x) \sim \mathcal{GP}\left(0, \mathcal{K}_{\alpha,\tau}(x, x')\right),
\end{equation*}
where the covariance kernel \( \mathcal{K}_{\alpha,\tau} \) is:
\begin{equation*}
    \mathcal{K}_{\alpha,\tau}(x, x') = (-\Delta + \tau^2)^{-\alpha}.
\end{equation*}
Here, \( \Delta \) represents the Laplacian operator.

The second scenario focuses on changes in the right-hand side term of the PDE while keeping the parameters \( \alpha \) and \( \tau \) constant. For the target dataset \( X^T \), we have a right-hand side term \( f = 1 \), and for the source dataset \( X^S \), the term is \( f = 5xy \). The PDEs are represented as:
\begin{equation*}
    \mathcal{L}_{\alpha,\tau}[u] = f_{X^T}, \quad \text{with} \quad f_{X^T} = 1,
\end{equation*}
for \( X^T \), and
\begin{equation*}
    \mathcal{L}_{\alpha,\tau}[u] = f_{X^S}, \quad \text{with} \quad f_{X^S} = 5xy,
\end{equation*}
for \( X^S \).

In these scenarios, \( \mathcal{L}_{\alpha,\tau} \) denotes the differential operator parametrized by \( \alpha \) and \( \tau \), and \( u \) signifies the solution to the PDE.

Each table lists the results for different quantities of $X^T$, specifically 20, 50, 100, and 500 samples, compared against the performance metrics of POD-DeepONet, standard DeepONet, and FNO. These comparisons (Table \ref{tab:darcy_target_dataset_1} and \ref{tab:darcy_target_dataset_2}) illustrate our framework's enhanced transfer learning capabilities, effectively showcasing its potential in handling diverse and challenging PDE scenarios with limited data.

\begin{table*}[h!]
\centering
\begin{tabular}{lcccc}
\toprule
$X^T$ Samples & POD-DeepONet & DeepONet & FNO & FF POD-DeepONet \\
\midrule
20 & 0.1914 & 0.2203 & 0.2187 & \textbf{0.1723} \\
50 & 0.1766 & 0.1850 & 0.1822 & \textbf{0.1504} \\
100 & 0.1015 & 0.1239 & 0.1117 & \textbf{0.0931} \\
500 & 0.0901 & 0.1065 & 0.1012 & \textbf{0.0854} \\
\bottomrule
\end{tabular}
\caption{Transfer learning performance on the target dataset $X^T$ with different data distribution.}
\label{tab:darcy_target_dataset_1}
\end{table*}

\begin{table*}[h!]
\centering
\begin{tabular}{lcccc}
\toprule
$X^T$ Samples & POD-DeepONet & DeepONet & FNO & FF POD-DeepONet \\
\midrule
20 & 0.3025 & 0.3198 & 0.3175 & \textbf{0.2074} \\
50 & 0.2102 & 0.2351 & 0.2242 & \textbf{0.1567} \\
100 & 0.1474 & 0.1422 & 0.1407 & \textbf{0.1216} \\
500 & 0.1121 & 0.1268 & 0.1219 & \textbf{0.0950} \\
\bottomrule
\end{tabular}
\caption{Transfer learning performance on the target dataset $X^T$ with different PDE terms.}
\label{tab:darcy_target_dataset_2}
\end{table*}

\subsection{Burgers' Equation}
\subsubsection{Background of Burgers' Equation}

We also consider the one-dimensional Burgers' equation as a test example for evaluating the performance of our architecture. The Burgers' equation is a fundamental partial differential equation from fluid mechanics. It is defined on the unit torus as follows:

\begin{equation}
\partial_t u(x, t) + \frac{1}{2} \partial_x (u(x, t)^2) = \nu \partial_{xx} u(x, t), \quad x \in (0, 1), t \in (0, 1)
\end{equation}
with the initial condition:
\begin{equation}
u(x, 0) = u_0(x), \quad x \in (0, 1).
\end{equation}

Here, \( u(x, t) \) represents the velocity field, \( \nu \) is the viscosity coefficient, and \( u_0(x) \) is the initial velocity distribution. In our operator learning framework, we aim to learn the mapping from the initial velocity distribution \( u_0(x) \) to the velocity field \( u(x, t) \) at subsequent times under the effects of advection and diffusion.

\subsubsection{Operator Learning}

Our operator learning framework aims to learn the mapping from the initial condition \(u_0(x)\) to the velocity field \(u(x,t)\) at future times, leveraging the capabilities of DeepONets enhanced by fusion frame theory to capture the nonlinear dynamics described by the Burgers' equation. 



Initially, we use FFN to derive 20 subspaces from the target dataset. Subsequently, POD decomposition is performed within each subspace, with an optimal selection of 30 POD modes per subspace.

To validate the effectiveness of our framework, we conducted a comprehensive comparison against the traditional POD-DeepONet, the standard DeepONet, and the Fourier Neural Operator (FNO) only in the source dataset. The evaluation demonstrates our framework's superiority in terms of predictive accuracy in Table \ref{tab:bur_source_dataset}.

\begin{table*}[h!]
\centering
\begin{tabular}{lcccc}
\toprule
Method & POD-DeepONet & DeepONet & FNO & FF POD-DeepONet \\
\midrule
MSE & 0.0174 & 0.0201 & 0.0182 & \textbf{0.0127} \\

\bottomrule
\end{tabular}
\caption{Performance comparison on the source dataset $X^S$.}
\label{tab:bur_source_dataset}
\end{table*}

\subsubsection{Transfer Operator Learning}

In the context of transfer learning, we examine the framework's adaptability in learning the Burgers' dynamics under viscosity coefficients, highlighting its robustness and accuracy in predicting complex, time-evolving patterns.

Furthermore, we retain the aforementioned subspaces and POD configurations while testing on the target dataset $X^T$. For the transfer operator learning, we consider one distinct scenario, the viscosity coefficient \( \nu \) is changed from 0.1 to 0.001.

In this scenario, for the target dataset \( X^T \), we set the \( \nu \) is 0.1, and for the source dataset \( X^S \), the \( \nu \) is 0.001. The PDEs are represented as:
\begin{equation*}
    \mathcal{L}_{\nu_{X^T}}[u] = f, \quad \text{with} \quad \nu = 0.1,
\end{equation*}
for \( X^T \), and
\begin{equation*}
    \mathcal{L}_{\nu_{X^S}}[u] = f, \quad \text{with} \quad \nu = 0.001,
\end{equation*}
for \( X^S \).

In these scenarios, \( \mathcal{L}_{\nu} \) denotes the differential operator parametrized by \( \nu \), and \( u \) signifies the solution to the PDE.

The table lists the results for different quantities of $X^T$, specifically 20, 50, 100, and 500 samples, compared against the performance metrics of POD-DeepONet, standard DeepONet, and FNO. These comparisons (Table \ref{tab:bur_target_dataset_1} illustrate our framework's enhanced transfer learning capabilities, effectively showcasing its potential in handling diverse and challenging PDE scenarios with limited data.

\begin{table*}[h!]
\centering
\begin{tabular}{lcccc}
\toprule
$X^T$ Samples & POD-DeepONet & DeepONet & FNO & FF POD-DeepONet \\
\midrule
20 & 0.0231 & 0.0236 & 0.0228 & \textbf{0.0197} \\
50 & 0.0217 & 0.0215 & 0.0202 & \textbf{0.0169} \\
100 & 0.0177 & 0.0202 & 0.0189 & \textbf{0.0130} \\
\bottomrule
\end{tabular}
\caption{Transfer learning performance on the target dataset $X^T$ with different $\nu$.}
\label{tab:bur_target_dataset_1}
\end{table*}

\subsection{Elasticity Model}

\subsubsection{Background of Elasticity Model}
The Elasticity Model is pivotal for analyzing material behavior when subjected to external forces. This model includes the equilibrium equation, relating the stress within a material to the body forces acting upon it. In the context of plane stress elasticity, applicable to thin plates, the governing PDE is:

\begin{equation}
\nabla \cdot \sigma(x) + f(x) = 0, \quad x = (x, y),
\label{eq:elasticity}
\end{equation}

where \( \sigma \) represents the stress tensor, and \( f(x) \) is the body force vector field. This equilibrium equation above is coupled with the constitutive relation for linear elastic materials under plane stress, given by:

\begin{equation}
\begin{bmatrix}
\sigma_{xx} \\
\sigma_{yy} \\
\tau_{xy} 
\end{bmatrix} = 
\frac{E}{1-\nu^2}
\begin{bmatrix}
1 & \nu & 0 \\
\nu & 1 & 0 \\
0 & 0 & \frac{1-\nu}{2}
\end{bmatrix}
\begin{bmatrix}
\frac{\partial u}{\partial x} \\
\frac{\partial v}{\partial y} \\
\frac{\partial u}{\partial y} + \frac{\partial v}{\partial x}
\end{bmatrix},
\label{eq:stress_strain}
\end{equation}

where \( E \) is Young's modulus, \( \nu \) is Poisson's ratio, \( \sigma_{xx} \) and \( \sigma_{yy} \) are normal stresses, and \( \tau_{xy} \) is the shear stress. The displacement field, described by \( u(x) \) and \( v(x) \), is the primary output of interest, representing the deformation of the material body under load.

The objective of the Elasticity Model in the operator learning context is to learn an operator that maps the input function \( f(x) \), encompassing the boundary conditions and varying material properties, to the output displacement fields \( u(x) \) and \( v(x) \). This mapping can be influenced by spatial variations in material properties, typically represented as a Gaussian random field:

\begin{equation}
\mathcal{X}(x, x') = \exp\left(-\frac{\lVert x - x' \rVert^2}{2l^2}\right),
\label{eq:gaussian_field}
\end{equation}

with correlation length \( l \) and variance parameters that embody the stochastic nature of material properties across the domain.

\subsubsection{Operator Learning}


For our proposed Fusion Frame POD-DeepONet model, initially, we employ a Fusion Frame Network (FFN) to derive subspaces from the target dataset $Y_u^S$ and $Y_v^S$, effectively partitioning the input space into 20 distinct subspaces. Subsequently, POD decomposition is performed within each subspace, with an optimal selection of 80 POD modes per subspace, as empirically validated through our testing. 


To validate the effectiveness of our framework, we conducted a comprehensive comparison against the traditional POD-DeepONet, the standard DeepONet, and the Fourier Neural Operator (FNO) only in the source dataset. The evaluation demonstrates our framework's superiority in terms of predictive accuracy in Table \ref{tab:source_dataset}.

\begin{table*}[h!]
\centering
\begin{tabular}{lcccccccc}
\toprule
Method & \multicolumn{2}{c}{POD-DeepONet} & \multicolumn{2}{c}{DeepONet} & \multicolumn{2}{c}{FNO} & \multicolumn{2}{c}{FF POD-DeepONet} \\
\midrule
Variable & \(u\) & \(v\) & \(u\) & \(v\) & \(u\) & \(v\) & \(u\) & \(v\) \\
\midrule
MSE & 0.0291 & 0.0303 & 0.0313 & 0.0322 & 0.0311 & 0.0318 & \textbf{0.0269} & \textbf{0.0272} \\
\bottomrule
\end{tabular}
\caption{Performance comparison on the source dataset \(X^S\) for variables \(u\) and \(v\).}
\label{tab:source_dataset}
\end{table*}

\subsubsection{Transfer Operator Learning}

We address the transfer learning scenarios in the context of the Elasticity Model under variation in external loading conditions. The transfer learning scenario of interest involves a variation in the sampling of the external force \( f \), which could represent different loading conditions in practical applications. Furthermore, we retain the aforementioned subspaces and POD configurations while testing on the target dataset $X^T$. 


In this scenario, for the target dataset \( X^T \), we set the \( l \) is 0.12, and for the source dataset \( X^S \), the \( l \) is 0.04.
The PDEs are represented as:
\begin{equation*}
    \mathcal{L}[u] + f_{l_{X^T}} = 0, \quad \text{with} \quad l = 0.12,
\end{equation*}
for \( X^T \), and
\begin{equation*}
    \mathcal{L}[u] + f_{l_{X^S}}= 0, \quad \text{with} \quad l = 0.04,
\end{equation*}
for \( X^S \).

In these scenarios, \( \mathcal{L} \) denotes the differential operator for different \( l \), and \( u \) signifies the solution to the PDE.

The table lists the results for different quantities of $X^T$, specifically 20, 50, 100, and 500 samples, compared against the performance metrics of POD-DeepONet, standard DeepONet, and FNO. These comparisons (Table \ref{tab:target_dataset_1}) illustrate our framework's enhanced transfer learning capabilities, effectively showcasing its potential in handling diverse and challenging PDE scenarios with limited data.

\begin{table*}[h!]
\centering
\begin{tabular}{lcccccccc}
\toprule
$X^T$ Samples & \multicolumn{2}{c}{POD-DeepONet} & \multicolumn{2}{c}{DeepONet} & \multicolumn{2}{c}{FNO} & \multicolumn{2}{c}{FF POD-DeepONet} \\
\midrule
 & \(u\) & \(v\) & \(u\) & \(v\) & \(u\) & \(v\) & \(u\) & \(v\) \\
\midrule
20 & 0.0621 & 0.0643 & 0.0703 & 0.0712 & 0.0651 & 0.0678 & \textbf{0.0588} & \textbf{0.0622} \\
50 & 0.0501 & 0.0506 & 0.0514 & 0.0518 & 0.0501 & 0.0517 & \textbf{0.0449} & \textbf{0.0457} \\
100 & 0.0302 & 0.0328 & 0.0320 & 0.0331 & 0.0330 & 0.0337 & \textbf{0.0290} & \textbf{0.0312} \\
\bottomrule
\end{tabular}
\caption{Transfer learning performance on the target dataset $X^T$ for variables $u$ and $v$ with different sample sizes.}
\label{tab:target_dataset_1}
\end{table*}


We compare the Fusion Frame POD-DeepONet framework's performance with traditional learning models through transfer learning scenarios involving external forces. The results demonstrate the framework's ability to generalize and efficiently transfer knowledge between varied conditions. 



\label{subsec:discussion}
The adaptability of the Fusion Frame POD-DeepONet to new conditions without extensive retraining underlines its potential for practical applications in computational material science. The ability to efficiently transfer knowledge between different material behaviors and loading conditions can significantly expedite the design and analysis of engineering structures.

\section{Related Work}

In the field of operator learning, the theories of Frame Theorem and Basis have provided powerful tools for bridging the gap between continuous and discrete representations. Frame theory is essential for understanding the relationships between functions and their discrete representations. Frame theory provides tools for reconstructing a function from its discrete samples. The Frame Theorem offers a mathematical tool for analyzing and handling the conversion issues between continuous and discrete representations encountered in operator learning. It focuses particularly on maintaining the consistency of operators across different resolutions. In contrast, Basis serves as a means for effective data representation and compression, playing a critical role in the field. \cite{2024frame} introduces the Representational Equivalence Neural Operator (ReNO) framework and showcases the application of the Frame Theorem to address the inconsistencies between continuous and discrete representations in operator learning, ensuring representational equivalence across varying discretizations and grid resolutions. Furthermore, the use of Principal Component Analysis (PCA) as a method of basis transformation indirectly highlights the value of Basis in operator learning by efficiently compressing data and preserving essential information, thus simplifying the learning process and enhancing computational efficiency\cite{de2022costPCAnet,bhattacharya2021model}. The Spectral Neural Operator (SNO) emphasizes the application of specific bases in operator learning through spectral methods, providing a robust tool for efficiently representing and processing functions in the frequency domain\cite{SNO}. The work BelNet \cite{zhang2023belnet} also uses the bases function to improve the ability of operator learning, thereby making operator learning more effective in handling complex samplings.

The Frame Theorem and Basis theory provide an effective methodological framework for operator learning, with the above applications collectively demonstrating their importance and practicality in advancing the field. While enhancing the performance of operator learning models, they also offer a theoretical foundation for the application of transfer learning. To provide a comprehensive overview of transfer learning algorithms, including basic concepts, main methods, and broad applications, one may refer to the review article by \cite{tan2018survey}. This introduction to basic methods and an in-depth exploration of transfer learning deepen our understanding of how to effectively leverage transfer learning algorithms to enhance operator learning performance.

The integration of transfer learning with deep operator networks (DeepONet) has shown promising potential in enhancing the performance of operator learning. Xu et al. cite{xu2023transfer} proposed a method that employs transfer learning to sequentially update DeepONets learned at different time frames. This approach enables the evolved DeepONets to more accurately track the changing complexity of evolutionary equations over time. Their method demonstrates significant improvements in long-term prediction accuracy, showcasing the potential of transfer learning in enhancing operator learning performance. In another vein, Wang et al. \cite{wang2023long} presented a strategy for adapting propagators learned through Physics-Informed DeepONet for specific evolutionary equations. By fine-tuning a small number of parameters in the final layer of the network without retraining the entire network, this method quickly achieves high-quality solutions to new differential equations. This approach alleviates the need for retraining neural networks when faced with different initial conditions, boundary conditions, or parameters for the same PDEs, thereby reducing computational burdens significantly. The work by Goswami et al. \cite{goswami2022tldon} introduces a transfer learning framework for enhancing DeepONet performance under conditional shifts. By leveraging a novel hybrid loss function within a task-specific learning context, this framework has remarkable efficiency and adaptability across diverse nonlinear PDE scenarios. These studies collectively highlight the latent power of applying transfer learning within the domain of operator learning. This introduces a new trajectory for development within the operator learning field, promising for advancing both theoretical understanding and practical applications.

Despite the widespread application of traditional transfer learning methods across various domains, from image recognition to natural language processing, these methods often fall short in precision or efficiency when addressing the unique application challenges of operator learning. This paper proposes a method based on the Frame Theorem aimed at resolving application issues within operator learning, introducing a transfer learning strategy better suited to the characteristics of operator learning. By combining these transfer learning methods with Frame Theorem, we provide a potential pathway for efficiently learning more complex and challenging tasks, thereby offering a new direction for the development of the operator learning field.

\section{Conclusion}
This paper introduced a novel approach to enhance the transferability of operator learning models by integrating fusion frame theory with the Proper Orthogonal Decomposition-enhanced Deep Operator Network (POD-DeepONet). Our innovative architecture leverages the robustness and flexibility of fusion frames to provide a more efficient and adaptable framework for solving Partial Differential Equations (PDEs) across various scientific and engineering domains. The experimental results demonstrate significant improvements in generalization capabilities across different PDE scenarios, affirming the effectiveness of our method.

The integration of fusion frame theory not only aids in improving traditional operator learning but also facilitates a deeper understanding of the underlying dynamics of PDEs, thereby enhancing the model's predictive performance. Future work will explore further optimizations of the fusion frame parameters and extend the framework to more complex multi-physics and multi-scale problems, potentially opening new avenues for research in computational science and engineering.

\bibliographystyle{alpha}
\bibliography{sample}

\end{document}